\documentclass[letterpaper]{article} % DO NOT CHANGE THIS
\usepackage{aaai25}  % DO NOT CHANGE THIS
\usepackage{times}  % DO NOT CHANGE THIS
\usepackage{helvet}  % DO NOT CHANGE THIS
\usepackage{courier}  % DO NOT CHANGE THIS
\usepackage[hyphens]{url}  % DO NOT CHANGE THIS
\usepackage{graphicx} % DO NOT CHANGE THIS
\usepackage{natbib}  % DO NOT CHANGE THIS AND DO NOT ADD ANY OPTIONS TO IT
\usepackage{caption} % DO NOT CHANGE THIS AND DO NOT ADD ANY OPTIONS TO IT
\usepackage{algorithm}
\usepackage{algorithmic}

\usepackage{newfloat}
\usepackage{listings}
\DeclareCaptionStyle{ruled}{labelfont=normalfont,labelsep=colon,strut=off} % DO NOT CHANGE THIS
\floatstyle{ruled}
\newfloat{listing}{tb}{lst}{}
\floatname{listing}{Listing}
\title{Global Graph Propagation with Hierarchical Information Transfer for Incomplete Contrastive Multi-view Clustering}
\author {
    Guoqing Chao\textsuperscript{\rm 1}\thanks{Corresponding authors, guoqingchao10@gmail.com},
    Kaixin Xu\textsuperscript{\rm 1},
    Xijiong Xie\textsuperscript{\rm 2},
    Yongyong Chen\textsuperscript{\rm 3}
}
\affiliations {
    \textsuperscript{\rm 1}Harbin Institute of Technology, Weihai\\
    \textsuperscript{\rm 2}Ningbo University\\
    \textsuperscript{\rm 3}Harbin Institute of Technology, Shenzhen
}

\usepackage{booktabs}
\usepackage{tabularx}
\newcolumntype{Z}{>{\centering\let\newline\\\arraybackslash\hspace{0pt}}X}
\usepackage{bm}
\usepackage{multirow}
\usepackage[graphicx]{realboxes}
\usepackage{amsmath}
\usepackage{amssymb}
\usepackage{makecell}
\usepackage{pifont}
\usepackage{epstopdf}
\usepackage{enumitem}
\usepackage{subfig}
\graphicspath{{image/}}

\begin{document}

 \maketitle

\begin{abstract}
Incomplete multi-view clustering has become one of the important research problems due to the extensive missing multi-view data in the real world. Although the existing methods have made great progress, there are still some problems: 1) most methods cannot effectively mine the information hidden in the missing data; 2) most methods typically divide representation learning and clustering into two separate stages, but this may affect the clustering performance as the clustering results directly depend on the learned representation. To address these problems, we propose a novel incomplete multi-view clustering method with hierarchical information transfer. Firstly, we design the view-specific Graph Convolutional Networks (GCN) to obtain the representation encoding the graph structure, which is then fused into the consensus representation. Secondly, considering that one layer of GCN transfers one-order neighbor node information, the global graph propagation with the consensus representation is proposed to handle the missing data and learn deep representation. Finally, we design a weight-sharing pseudo-classifier with contrastive learning to obtain an end-to-end framework that combines view-specific representation learning, global graph propagation with hierarchical information transfer, and contrastive clustering for joint optimization. Extensive experiments conducted on several commonly-used datasets demonstrate the effectiveness and superiority of our method in comparison with other state-of-the-art approaches. The code is available at https://github.com/KelvinXuu/GHICMC.

\end{abstract}

\section{Introduction}
Multi-view data refers to the same data objects observed from different viewpoints. These views provide diverse knowledge about the same objects, which are generally complementary and consistent. Multi-view learning~\cite{yan2021deep} is a popular machine learning paradigm to mine the complementary and consistent information within multi-view data, and multi-view clustering is the unsupervised branch of multi-view learning. Multi-view clustering aims to improve the clustering performance by exploiting the information within multiple views.

Most existing multi-view clustering methods generally rely on the samples with complete view data~\cite{chao2022incomplete, xu2022self}. However, this assumption is not established in many real-world scenarios since some samples may only have incomplete views owing to sensor instability or other unavoidable factors. Therefore, incomplete multi-view clustering (IMVC) has attracted increasing attention~\cite{chao2021survey}.

To deal with incomplete multi-view data problems, many traditional incomplete multi-view clustering methods have achieved great performance~\cite{wang2020deep}. These methods can be divided into three classes: matrix factorization based method~\cite{hu2019one}, kernel learning based method~\cite{liu2019multiple, liu2019efficient}, and graph learning based method~\cite{wen2019unified}. Compared with recent deep learning based methods, these traditional approaches possess certain interpretability but limited representation capability. In addition, commonly utilized singular value decomposition and matrix inverse operations increase time complexity.

In recent years, many deep learning methods~\cite{lin2022dual, yu2024deep, chao2024incomplete} sprung up. Some imputation-based deep learning clustering methods, for example, the generative adversarial network (GAN) is used to recover the missing data and then perform traditional multi-view clustering~\cite{wang2018partial}. However, there is no rigorous theory to guarantee the accuracy of the recovered data, especially in high missing rates scenarios. The imputation-free IMVC obtain the final clustering results from the available views directly~\cite{hu2019one, xu2021multi, xu2022deep}. The vast majority of these methods obtain a unified representation by mapping the individual view data to a latent consensus space and fusing the multi-view representation to perform traditional clustering methods finally~\cite{li2021incomplete}. Obviously, the imputation-free approaches cannot effectively use the information hidden in the missing data. Since most methods divide representation learning and clustering into two separate stages, the final clustering performance will be suboptimal.

In this paper, we propose Global graph propagation with Hierarchical information transfer for Incomplete Contrastive Multi-view Clustering (GHICMC) to address the above problems. Specifically, we use GCN to obtain the embedded representations, which can capture the topological structure. Further, we integrate the view-specific features into a consensus representation and use the global graph to transfer information in the same layer of the graph neural networks. In this way, missing data can be imputed adaptively in each view under both intra-view and inter-view relations. In contrast to most of the existing methods, we design a weight-shared pseudo-classifier to obtain the final clustering results, using contrastive learning to obtain the distinguishable clustering results, which can achieve end-to-end optimization.

The major contributions are summarized as follows:

\begin{itemize}
    \item We propose a novel incomplete multi-view clustering method GHICMC, which can unify representation learning, missing data handling, and clustering into a joint framework for optimization.
    \item We propose a novel global graph propagation module that can impute multi-view data adaptively with hierarchical information transfer in graph convolutional layers, which can not only effectively use the local graph structure of each view but also learn the global graph structure across views.
    \item The experiment results on five multi-view datasets demonstrate the effectiveness and superiority of our GHICMC over several state-of-the-art methods.
\end{itemize}

\section{Related Works}
\subsection{Incomplete Multi-view Clustering}
Incomplete multi-view clustering can be roughly classified into two categories: imputation-based methods and imputation-free methods. For imputation-based IMVC methods, the missing data is often recovered as the first step, followed by the usage of classic multi-view clustering methods for clustering.~\citeauthor{wang2021generative} use GAN to recover missing data and fuse features across views using adaptive fusion mechanism~\cite{wang2021generative}.~\citeauthor{li2023incomplete} use prototypes and sample-prototype relationships to recover missing samples, which preserve sample commonality and view diversity~\cite{li2023incomplete}.~\citeauthor{wang2024joint} use graph topological structure to recover the missing data and learn distinctive features by graph contrastive learning~\cite{wang2024joint}. For imputation-free IMVC methods, they usually perform clustering operations directly based on the available view data. For instance,~\citeauthor{xu2022deep} project embedded representations of complete data to a higher dimensional space for linear separability identification~\cite{xu2022deep}.~\citeauthor{xu2023adaptive} project all available data into a consensus feature space and explore consensus cluster information by maximizing mutual information~\cite{xu2023adaptive}.~\citeauthor{xu2024deep} use variational autoencoders to extract the information of each view, and use Product-of-Experts to obtain a consensus representation~\cite{xu2024deep}.

\subsection{Graph-based Incomplete Multi-view Clustering}
While many methods focus on learning high-level representations of data, the structural relations of the data are largely ignored. Since the graph contains the topological relations between the data and can better represent the structural relations between the data, graph-based IMVC method has also become an effective IMVC method.~\citeauthor{li2021incomplete} use a consensus graph to obtain clustering results by adaptively weighting the stretched base partition~\cite{li2021incomplete}. Instead of imputing missing samples directly,~\citeauthor{liu2023self} impute missing instances by completing the similarity graph for a clustering task~\cite{liu2023self}. With the development of deep learning technology, GCN-based IMVC methods are also widely used, which can capture the structure of data. SDIMC-net~\cite{wen2021structural} uses view-specific GCN encoders to simultaneously mine high-level representations and higher-order structural information of data.~\citeauthor{wang2022incomplete} propose a module for completing cross-view relationship transfer, which transfers the similarity between existing samples to missing samples and utilizes GCN to impute missing data~\cite{wang2022incomplete}. ~\citeauthor{yang2024geometric} propose the weight assignment method in a geometric perspective for the graph aggregation algorithm~\cite{yang2024geometric}.

\section{The Proposed Method}
\label{method}
\begin{figure*}[t]
    \centering
    \includegraphics[width=0.98\linewidth]{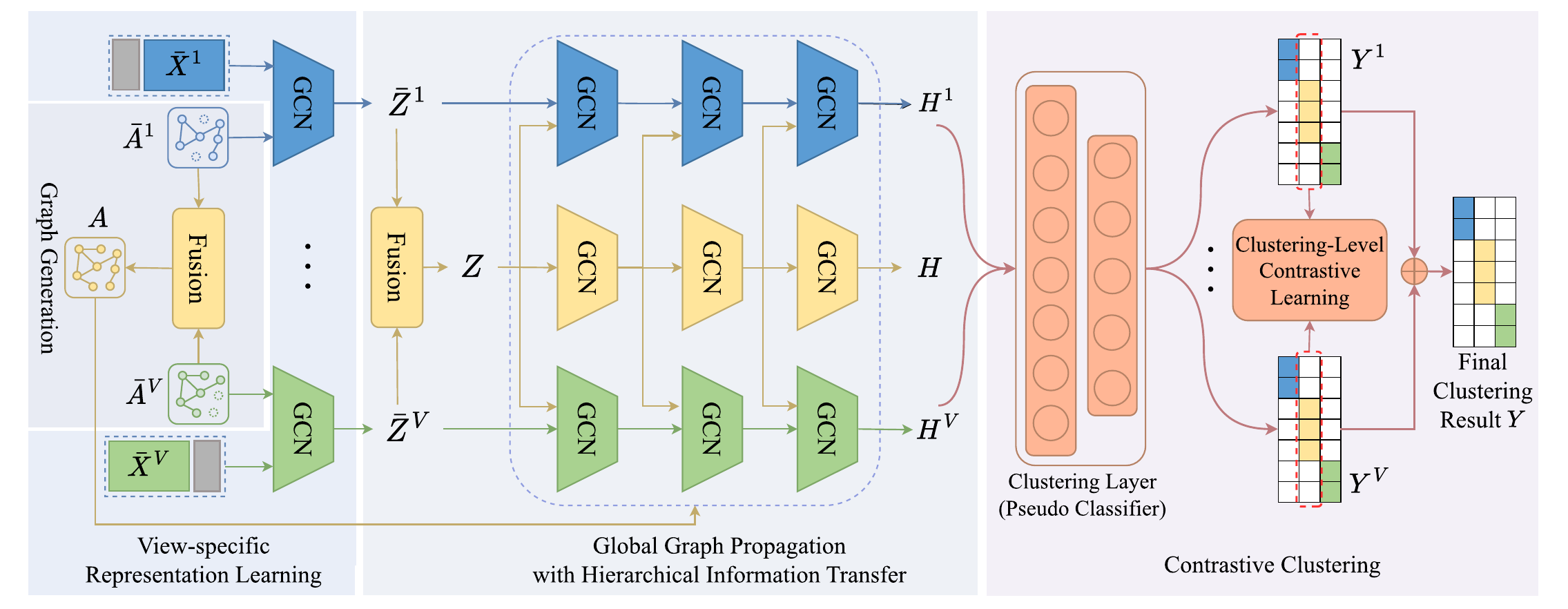}
    \caption{The overall framework of GHICMC. It consists of three main modules: view-specific representation learning, global graph propagation with hierarchical information transfer, and a contrastive clustering module. Firstly, the view-specific GCN is employed to learn the view-specific representation $\{\bm{\bar{Z}}^v\}^V_{v=1}$ with the local graph, which is then fused to obtain the consensus representation $\bm{Z}$. Secondly, a global graph propagation module with hierarchical information transfer is used to impute the representations adaptively and simultaneously learn deeper representations $\{\bm{H}^v\}^V_{v=1}$ with the consensus representation $\bm{Z}$ and global graph $\bm{A}$ which is constructed by $\{\bm{\bar{A}}^v\}^V_{v=1}$. Finally, the weight-sharing pseudo-classifier is employed to generate the clustering predictions, from which we derive the final clustering result.}
    \label{fig:model}
\end{figure*}

In this section, we provide a thorough introduction to the proposed method which is composed of three main joint learning modules, namely, view-specific representation learning module, global graph propagation module with hierarchical information transfer, and contrastive clustering module. The overall architecture of the proposed method is shown in Figure~\ref{fig:model}.

\subsubsection{Notations}
Formally, we define a multi-view dataset $\bm{X}=\{ \bm{X}^1, \bm{X}^2, \cdots,\bm{X}^V \}$, where $V$ is the number of views. The feature matrix of the $v$-th view, denoted as $\bm{X}^v=\{\bm{x}_1^v, \bm{x}_2^v, \cdots, \bm{x}_{N}^v \}\in \mathbb{R}^{N\times d_v}$, is a set of $N$ samples with the feature dimension $d_v$. Let $n_v$ denotes the number of existent samples in view $v$ and $\bm{\bar{X}}^v \in \mathbb{R}^{n_v\times d_v}$ represents the complete sample in the $v$-th view. We define $\bm{M}\in \mathbb{R}^{N\times V}$ to indicate the presence or absence of data entry where $\bm{M}_{ij}=1$ means the $i$-th sample exists in the $j$-th view, otherwise $\bm{M}_{ij}=0$. In addition, we use $C$ to represent the total number of clusters.

\subsection{Graph Generation Module}
For each view, we define the indicator vector $h^v \in \mathbb{R}^{n_v}$ which indicates the sequence numbers of $n_v$ sorted complete data, where for any $i$, $M_{h^v_iv}=1$. Then the indicator matrix $\bm{F}^v \in {\{0,1\}}^{n_v\times N}$ can be calculated as follows:
\begin{equation}\label{indicator_M}
	\bm{F}_{ij}^v=\begin{cases}
	1,&\text{if $h_i^v=j$}\\
	0,&\text{otherwise.}
	\end{cases}
\end{equation}
Obviously, we can obtain the complete data matrix $\bm{\bar{X}}^v \in \mathbb{R}^{n_v\times d_v}$ which represents the samples that exist in the $v$-th view as follows:
\begin{equation}\label{X_exist}
    \bm{\bar{X}}^v=\bm{F}^v \times \bm{X}^v.
\end{equation}
In order to generate graphs with complete data, the similarity matrix $\bm{\bar{S}}^v \in \mathbb{R}^{n_v \times n_v}$ is first constructed from the existing samples of each view based on the Gaussian kernel:
$\bm{\bar{S}}_{ij}^v=e^{- \frac{{\left\| \bm{\bar{x}}_i^v-\bm{\bar{x}}_j^v \right\|}^2}{2\sigma^2} }$, where $\sigma$ is the bandwidth. After that, we utilize the K-nearest neighbors to construct the graph structure $\bm{\bar{A}}^v \in \mathbb{R}^{n_v \times n_v}$ based on the similarity matrix $\bm{\bar{S}}^v$.

We expand $\bm{\bar{A}}^v$ to obtain the graph structure $\bm{A}^v$ between the full data by setting the relation to 0 where the sample is absent. The specific operation is described as follows:
\begin{equation}\label{S_v}
    \bm{A}^v=\left(\bm{F}^v\right)^T \times \bm{\bar{A}}^v \times \bm{F}^v,
\end{equation}
where $\bm{A}^v \in \mathbb{R}^{N \times N}$ is the missing graph structure on the $v$-th view. The graphs of all views are fused into a global graph $\bm{A}$ in the following way:
\begin{equation}\label{eq:A_f}
    \bm{A}=\sum\limits_{v=1}^{V}{\pi^v \bm{A}^v},
\end{equation}
where ${\pi }^{(v)}$ is a parameter that can be learned to adjust the significance of the graph in the $v$-th view, which is initialized as $1/V$. Moreover, ${\pi }^{(v)}$ is normalized as:
\begin{equation}
  {{\pi }^{(v)}}\leftarrow \frac{{{e}^{{{\pi }^{(v)}}}}}{\sum\nolimits_{k=1}^{V}{{{e}^{{{\pi }^{(k)}}}}}}.
\end{equation}

\subsection{View-Specific Representation Learning with Local Graph}
The graph auto-encoder is capable of learning structural information as well as capturing node feature information. In order to obtain a useful representation, the view-specific GCN encoder is employed for each of the $V$ views. Specifically, the existing samples $\bm{\bar{X}}^v$ and the corresponding view-specific graph $\bm{\bar{A}}^v$ are fed into the view-specific GCN encoder to obtain the view-specific representation, as follows:
 \begin{equation}\label{H_v}
    \bm{\bar{Z}}^v=f_v\left(\bm{\bar{X}}^v,\bm{\bar{A}}^v;\theta^v\right),
\end{equation}
where $\bm{\bar{Z}}^v\in \mathbb{R}^{n_v \times d_v}$ represents the view-specific representations of the existing samples in $v$-th view, $f_v$ is the GCN encoder of view $v$ and $\theta^v$ is the corresponding parameter.
The graph convolution operation of the $l$-th layer is represented as
\begin{equation}\label{eq:Z_vs}
    \bm{\bar{Z}}_{(l)}^v=\sigma \left(\bm{\hat{A}}^v\bm{\bar{Z}}_{(l-1)}^v \bm{W}_{(l)}^v\right),
\end{equation}
where $\bm{\hat{A}}^v=\left(\bm{\Tilde{D}}^v\right)^{-\frac{1}{2}}\bm{\Tilde{A}}^v\left(\bm{\Tilde{D}}^v\right)^{-\frac{1}{2}}$, $\Tilde{\bm{A}^v}=\bm{\bar{A}}^v+\bm{I}_{n_v}$ and $\Tilde{\bm{D}^v}_{ii}=\sum_j \Tilde{\bm{A}^v}_{ij}$.
$\bm{I}_{n_v}$ is the identity matrix that learns self-information, $\bm{W}_{(l)}^v$ is the trainable parameters in the $l$-th layer of the encoder for the $v$-th view, and $\sigma$ denotes the activation function like Relu, Sigmoid, and so on. The initial input $\bm{\bar{Z}}_{(0)}^v=\bm{\bar{X}}^v$. Through the propagation of the local graph, the view-specific representation can maintain the view-specific structure. 
After obtaining the view-specific latent representation $\bm{\bar{Z}}^v$, we use the view-specific decoder to reconstruct the multi-view data as follows:
\begin{equation}\label{eq:X_pre}
    {\bm{\hat{X}}^v=g_v\left(\bm{\bar{Z}}^v;\theta_g^v\right)},
\end{equation}
where $\bm{\hat{X}}^v \in \mathbb{R}^{n_v\times d_v}$ is the reconstructed feature, $g_v$ is the decoder composed of fully-connected layers for the $v$-th view and $\theta_g^v$ is the corresponding parameter of $v$-th decoder. The reconstruction loss $\mathcal{L}_{rec}$ between original data and reconstructed data can be calculated as follows:
\begin{equation}\label{eq:L_rec}
  \mathcal{L}_{rec}=\frac{1}{V}\sum\limits_{v=1}^{V}{\frac{1}{n_v}\left\| \bm{\hat{X}}^{v}-{{\bm{\bar{X}}}^{v}} \right\|_{F}^{2}}.
\end{equation}

\subsection{Global Graph Propagation with Hierarchical Information Transfer}
To address the challenge brought by missing data, most methods simply use the mean or mode of feature value to fill in the missing data. However, in most cases, this strategy doesn't work well and sometimes even degrades the performance.
Since GCN can aggregate the features of neighbor nodes to update its own features, we adopt GCN to resolve this problem. Specifically, we first obtain the consensus representation $\bm{Z}\in \mathbb{R}^{N \times d_v}$ by expanding $\bm{\bar{Z}}^v$ and then linearly combining them, as follows:
\begin{equation}\label{eq:Z_f}
    {\bm{Z}_{i}}=\sum\limits_{v=1}^{V}{\bm{Z}_i^v / \bm{T}_i },
\end{equation}
where $\bm{Z}^v=\left(\bm{F}^v\right)^T\bm{\bar{Z}}^{v}$ denotes the expanded representation which fills the missing value of each view with 0, and $\bm{T}_i=\sum\nolimits_{j=1}^{V}{{{\bm{M}}_{ij}}}$ represents the number of views where the $i$-th sample exists. Then the consensus representation $\bm{Z}$ is fed into the stacked GCN to obtain the representation $\bm{H}$ as follows:
\begin{equation}\label{eq:h}
    \bm{H}_{(l)}=\sigma \left(\bm{\hat{A}}\bm{H}_{(l-1)} \bm{W}_{(l)}\right),
\end{equation}
where $\bm{\hat{A}}=\bm{\Tilde{D}}^{-\frac{1}{2}}\bm{\Tilde{A}}\bm{\Tilde{D}}^{-\frac{1}{2}}$, $\Tilde{\bm{A}}=\bm{A}+\bm{I}_n$ and $\Tilde{\bm{D}}_{ii}=\sum_j \Tilde{\bm{A}}_{ij}$. Note that here $\bm{H}_{(0)}=\bm{Z}$, the output $\bm{H}_{(l)}$ of the $l$-th layer represents the aggregation of the $l$-order neighbor node features, the final output is represented as $\bm{H}$. Inspired by the fact that each convolutional layer of GCN aggregates one-order neighbor node information~\cite{li2019deepgcns}, we attempt to take the output of each convolutional layer of the consensus GCN as input and pass it through the same-layer convolutional layers of the other $v$ view-specific GCN. In this way, each incomplete multi-view data can perform representation imputation based on both local relations within the view and global relations across multiple views simultaneously through GCN's message-passing mechanism. Since the information is transferred in the same-order, the imputation of features will be more accurate. The operation of each layer in the $v$-th view is represented as follows:
\begin{equation}\label{eq:h_v}
    \bm{H}_{(l)}^v=\sigma \left(\bm{\hat{A}} \cdot \frac{1}{2}\left(\bm{H}_{(l-1)}^v+\bm{H}_{(l-1)}\right) \cdot \bm{W}_{(l)}^v\right),
\end{equation}
where $\bm{H}_{(0)}^v=\bm{Z}^v$. After the above information transfer, missing latent representation can be imputed based on view-specific representation and consensus representation simultaneously. We denote that $\bm{H}^v$ is the representation after the $v$-th view-specific GCN network with global graph propagation.

To make full use of the consistency across all the views, we constrain the distribution $\bm{P}^{v}$ of the $v$-th imputed representation with the distribution $\bm{Q}$ of the consensus representation. The distribution $\bm{P}^{v}$ and the distribution $\bm{Q}$ can be calculated respectively as follows:
\begin{equation}\label{eq:p_distrbution}
    \bm{P}_{ij}^{v}=\frac{{{\left( 1+{{\left\| \bm{H}_{i}^{v}-\bm{H}_{j}^{v} \right\|}^{2}} \right)}^{-1}}}{\sum\nolimits_{k\ne l}{{{\left( 1+{{\left\| \bm{H}_{k}^{v}-\bm{H}_{l}^{v} \right\|}^{2}} \right)}^{-1}}}},
\end{equation}
\begin{equation}\label{eq:q_distrbution}
    \bm{Q}_{ij}=\frac{{{\left( 1+{{\left\| \bm{Z}_{i}-\bm{Z}_{j} \right\|}^{2}} \right)}^{-1}}}{\sum\nolimits_{k\ne l}{{{\left( 1+{{\left\| \bm{Z}_{k}-\bm{Z}_{l} \right\|}^{2}} \right)}^{-1}}}}.
\end{equation}
The loss function $\mathcal{L}_{sc}$ is the KL divergence between the distribution of the imputed representation and the consensus representation which can be calculated as follows:
\begin{equation}\label{eq:L_sc}
    \mathcal{L}_{sc} = \sum_{v = 1}^{V}{KL\left(\bm{Q}\|\bm{P}^v\right)} = \sum_{v = 1}^{V}\sum\limits_{i} \sum\limits_{j} \bm{q}_{ij}\log\frac{\bm{q}_{ij}}{\bm{p}^v_{ij}}.
\end{equation}

\begin{algorithm}[ht]
	\caption{\textbf{GHICMC}}
	\label{alg:algm}
	\textbf{Input}: Multi-view dataset $\{\bm{X}^v\}^V_{v=1}$ with $n$ samples and $V$ views, cluster number $C$, initial neighbor number $K$, temperature parameter $\tau$, parameters $\alpha$, $\beta$, maximum iterations $epochs$\\
	\textbf{Output}: Clustering result $\bm{Y}$
	\begin{algorithmic}[1] 
		\STATE Compute $\{\bm{\bar{A}}^v\}^V_{v=1}$, $\{\bm{A}^v\}^V_{v=1}$ and $\bm{A}$ by Eq~\eqref{eq:A_f}
            \FOR{epoch = 1 to $epochs$} 
                \STATE Compute $\{\bm{\bar{Z}}^v\}^V_{v=1}$ by Eq~\eqref{eq:Z_vs}
                \STATE Compute consensus representation $\bm{Z}$ by Eq~\eqref{eq:Z_f}
                \STATE Compute $\{\bm{\hat{X}}^v\}^V_{v=1}$ by Eq~\eqref{eq:X_pre}
                \STATE Calculate the loss $\mathcal{L}_{rec}$ by Eq~\eqref{eq:L_rec}
                \STATE Compute representation $\bm{H}$ by Eq~\eqref{eq:h}
                \STATE Compute imputed representation $\bm{H}^v$ by Eq~\eqref{eq:h_v}
                \STATE Calculate the loss $\mathcal{L}_{sc}$ by Eq~\eqref{eq:L_sc}
                \STATE Compute soft assignment $\{\bm{Y}^v\}^V_{v=1}$ by Eq~\eqref{eq:Y}
                \STATE Calculate the loss $\mathcal{L}_{ccl}$ by Eq~\eqref{eq:L_ccl}
                \STATE Calculate the overall loss $\mathcal{L}$ by Eq~\eqref{eq:L_all}
                \STATE Update through gradient descent to minimize $\mathcal{L}$
		\ENDFOR
		\STATE Obtain the final clustering result $\bm{Y}$ by Eq~\eqref{eq:y_final}
		\RETURN $\bm{Y}$
	\end{algorithmic}
\end{algorithm}
\subsection{Contrastive Clustering Module}
In order to achieve end-to-end joint optimization, a weight-shared pseudo-classifier implemented with two fully connected layers is employed. Specifically, the clustering result  from the $v$-th view is obtained as follows:
\begin{equation}\label{eq:Y}
  \bm{Y}^v=\mbox{softmax}\left(\mbox{Relu}\left(\bm{H}^v \cdot \bm{W}_{(0)}+\bm{b}_{(0)}\right) \cdot \bm{W}_{(1)}+\bm{b}_{(1)}\right),
\end{equation}
where $\bm{Y}^v \in \mathbb{R}^{N \times C}$ is the soft clustering assignment as well as the representation for the $v$-th view and $C$ is the number of clusters. Each column represents the probability distribution for each sample contained in each cluster where $\bm{y}_j^v$ represents the $j$-th column of $\bm{Y}^v$:
\begin{equation}
    \bm{y}_j^v=
    \left[\begin{array}{c}
        \bm{Y}_{1j}^v\\
        ...\\
        \bm{Y}_{Nj}^v\\
    \end{array}\right].
\end{equation}
Then the cluster-level constrastive loss for the $j$-th sample between view $v$ and $w$ is formulated as:
\begin{equation}
  \mathcal{L}_{j}^{(vw)}=-\log \frac{{{e}^{d(y_{j}^{v},y_{j}^{w})/\tau }}}{\sum\nolimits_{u=v,w}\sum\nolimits_{k=1}^{C}{{{e}^{d(y_{j}^{v},y_{k}^{u})/\tau }}-{{e}^{\tfrac{1}{\tau }}}}},
\end{equation}
where $d(y_{j}^{v},y_{k}^{w})$ is the cosine distance to indicate the similarity between $y_{j}^{v}$ and $y_{k}^{w}$, and $\tau$ denotes the temperature parameter. Therefore, the cluster-level contrastive loss across multiple views is designed as:
\begin{equation}\label{eq:L_ccl}
  {\mathcal{L}_{ccl}}=\frac{1}{2C}\sum\limits_{j=1}^{C}{\sum\limits_{v=1}^{V}{\sum\limits_{w \neq v}^{}{\mathcal{L}_{j}^{(vw)}}}}+\sum\limits_{j=1}^{C}{\sum\limits_{v=1}^{V}{s_{j}^{v}\log s_{j}^{v}}},
\end{equation}
where $s_{j}^{v}$ is defined as $s_{j}^{v}=\frac{1}{N}\sum\nolimits_{i=1}^{N}{y_{ij}^{v}}$. The first term of Eq~\eqref{eq:L_ccl} aims to obtain the consistency across all the views.
The second term of Eq~\eqref{eq:L_ccl} is a regularization term to prevent the clustering of all samples into a single cluster~\cite{huang2020deep}.
After that, the final clustering result $\bm{Y}=\left[\bm{y}_1,\bm{y}_2,\cdots,\bm{y}_N\right]$ is obtained from all the $V$ views as follows:
\begin{equation}\label{eq:y_final}
    \bm{y}_i=\mathop{\arg\max}_{j}{\left(\frac{1}{V}\sum_{v=1}^{V}{\bm{Y}^v_{ij}} \right)}.
\end{equation}

\subsection{The Overall Loss Function}
In summary, we have introduced a novel method using global graph propagation with hierarchical information transfer for incomplete multi-view clustering. In the training stage, the view-specific representation learning module, global graph propagation module with hierarchical information transfer, and the contrastive clustering module are jointly optimized, and the total objective function is represented as follows:
\begin{equation}\label{eq:L_all}
{\mathcal{L}=\mathcal{L}_{rec}+\alpha\mathcal{L}_{sc}+\beta\mathcal{L}_{ccl}},
\end{equation}
where $\alpha$ and $\beta$ are the trade-off parameters to balance the different loss functions. The whole learning process of our model is summarized in Algorithm~\ref{alg:algm}.

\section{Experiments}

\subsection{Experimental Settings}
We adopt Adam to optimize the loss function, and all the experiments are conducted in PyTorch 2.1.0 on Ubuntu 20.04 with an NVIDIA 3060 GPU.
The hyper-parameter $K$ is set to 10. The temperature parameter $\tau$ is fixed at 0.5. The parameters $\alpha$ and $\beta$ are both set to 10.

To simulate incomplete multi-view datasets, we define the incomplete rate  $\eta=(N-m)/N$ as the percentage of samples that have missing views, where $N$ is the total number of samples and $m$ is the number of complete samples. For each incomplete sample, we randomly delete its views.

\begin{table}[!htbp]
\centering
\small
\tabcolsep=1.0mm
\begin{tabular}{cccc}
\toprule
Dataset         & $\#$Samples   &$\#$Clusters & $\#$Dimensions \\
\midrule
HandWritten     & 2000      & 10    & 240/76/216/64   \\
MSRC-V1         & 210       & 7     & 1302/48/512/100/256/210    \\
Scene-15        & 4485      & 15    & 20/59  \\
LandUse-21      & 2100      & 21    & 59/40  \\
100leaves       & 1600      & 100   & 64/64/64  \\
\bottomrule
\end{tabular}
\caption{The statistics of five datasets.}
\label{tab:dataset}
\end{table}

\subsection{Datasets and Evaluation Metrics}
We adopt five commonly-used datasets to evaluate our proposed model, the statistics of these datasets are illustrated in Table~\ref{tab:dataset}.

\textbf{HandWritten}~\cite{li2015large}:
It contains 2000 samples which are HandWritten digits 0-9. 
Four categories of features are selected as 4 views in our experiments.

\textbf{MSRC-V1}~\cite{winn2005locus}:
This original dataset has 240 images divided into eight classes, each with 30 images. We selected 210 photographs from seven categories. Six classes of visual characteristics are selected as six views. 

\textbf{Scene-15}~\cite{mallah2013plant}:
It is composed of 4485 images from 15 scene categories, with PHOG and GIST as two views.

\textbf{LandUse-21}~\cite{yang2010bag}:
It has 2100 satellite images distributed in 21 categories with two views: PHOG and LBP.

\textbf{100leaves}~\cite{mallah2013plant}:
It consists of 1600 samples from 100 categories. Shape descriptor, fine scale margin, and texture histogram features are used as three views.

Three clustering evaluation metrics: accuracy (ACC), normalized mutual information (NMI), and adjusted rand index (ARI) are adopted to evaluate the clustering performance. All these metrics range from 0 to 1, and a higher value of the metrics indicates a better clustering performance.

\begin{table*}[htb]
\centering
\tabcolsep=1.0mm
\small
    \begin{tabular}{clccccccccccccccc} \hline
    \multirow{2}{*}{\makecell{$\eta$}} & 
    \multirow{2}{*}{Method} & 
    \multicolumn{3}{c}{100leaves} & 
    \multicolumn{3}{c}{HandWritten} & 
    \multicolumn{3}{c}{LandUse-21} & 
    \multicolumn{3}{c}{MSRC-V1} &
    \multicolumn{3}{c}{Scene-15} \\
    &   
    & ACC & NMI & ARI 
    & ACC & NMI & ARI 
    & ACC & NMI & ARI 
    & ACC & NMI & ARI 
    & ACC & NMI & ARI \\ 
    \hline
    \multirow{9}{*}{0}
        & BSV 
        & 57.10 & 78.97 & 45.27 
        & 70.83 & 68.27 & 58.70
        & 23.37 & 26.89 & 9.02 
        & 70.76 & 62.53 & 51.94 
        & 36.28 & 38.45 & 20.80 \\
        
        & Concat 
        & 73.08 & 89.26 & 67.47 
        & 80.37 & 78.71 & 71.89 
        & 22.63 & 27.85 & 9.60 
        & 73.24 & 70.31 & 60.80 
        & 38.47 & 40.87 & 22.34 \\
        
        & DAIMC 
        & 68.83 & 86.86 & 61.95 
        & 79.40 & 75.28 & 68.69 
        & 26.95 & 30.56 & 12.53 
        & 88.76 & 80.94 & $\underline{77.54}$
        & $\underline{39.89}$ & $\underline{42.78}$ & $\underline{24.53}$ \\
        
        & UEAF 
        & 34.77 & 66.89 & 22.15 
        & 61.12 & 60.33 & 47.21 
        & 21.30 & 27.32 & 7.93 
        & 54.57 & 45.58 & 29.58 
        & 34.92 & 37.17 & 18.76 \\
        
        & DCP 
        & 75.71 & 90.11 & 68.58 
        & 84.60 & 86.51 & 79.22 
        & 27.11 & 31.89 & $\underline{14.27}$
        & 86.38 & 78.92 & 73.27 
        & 38.70 & 42.54 & 22.49 \\
        
        &ANIMC 
        & 69.83 & 87.00 & 61.73 
        & 84.72 & 78.52 & 73.40 
        & $\underline{27.46}$ & $\underline{32.86}$ & 13.42 
        & 79.05 & 71.45 & 64.20 
        & 36.17 & 36.95 & 19.14 \\
        
        & LSIMVC 
        & 37.69 & 66.00 & 23.86 
        & 94.35 & 89.06 & 88.07 
        & 24.98 & 27.99 & 9.12 
        & 87.81 & 80.99 & 72.03 
        & 34.34 & 37.18 & 18.63 \\
        
        & PIMVC 
        & $\underline{83.31}$ & $\underline{93.88}$ & $\underline{79.34}$ 
        & $\underline{95.80}$ & $\underline{91.46}$ & $\underline{90.89}$ 
        & 19.62 & 28.11 & 7.28 
        & $\underline{90.00}$ & $\underline{83.35}$ & 77.28 
        & 35.99 & 39.43 & 20.80 \\
        
        & Ours 
        & $\bm{93.85}$ & $\bm{96.82}$ & $\bm{90.95}$ 
        & $\bm{97.44}$ & $\bm{94.16}$ & $\bm{94.40}$ 
        & $\bm{28.27}$ & $\bm{33.01}$ & $\bm{14.37}$ 
        & $\bm{95.14}$ & $\bm{90.64}$ & $\bm{88.62}$ 
        & $\bm{42.61}$ & $\bm{43.82}$ & $\bm{25.56}$ \\
        
        \hline
        \multirow{9}{*}{\makecell{0.5}}
        
        & BSV 
        & 42.69 & 59.76 & 6.76  
        & 55.92 & 49.54 & 27.19  
        & 18.65 & 19.36 & 3.28  
        & 55.81 & 45.32 & 26.61  
        & 28.54 & 27.60 & 8.49 \\
        & Concat 
        & 48.85 & 68.55 & 23.80  
        & 61.69 & 55.56 & 35.15  
        & 18.71 & 20.46 & 5.63  
        & 47.43 & 40.37 & 23.78  
        & 28.47 & 32.15 & 14.47 \\ 
        
        & DAIMC 
        & 42.02 & 66.51 & 27.61  
        & 76.59 & 72.17 & 65.51  
        & 20.76 & 21.45 & 7.96  
        & 78.67 & 70.78 & 63.58  
        & 32.99 & 34.48 & 18.01 \\ 
        
        & UEAF  
        & 27.18 & 56.04 & 9.67  
        & 41.93 & 40.36 & 25.78  
        & 16.64 & 18.51 & 3.70  
        & 58.95 & 48.99 & 34.06  
        & 28.38 & 28.01 & 8.19 \\
        
        & DCP 
        & 41.00 & 67.20 & 19.65  
        & 82.70 & 82.72 & 74.54  
        & $\underline{25.51}$ & $\underline{28.67}$ & $\bm{12.60}$  
        & 62.00 & 53.59 & 40.16  
        & $\underline{39.78}$ & $\underline{41.42}$ & $\underline{21.84}$ \\ 
         
        & ANIMC 
        & 46.09 & 69.50 & 28.45 
        & 85.53 & 75.07 & 71.45 
        & 18.71 & 20.57 & 6.34 
        & 74.29 & 62.18 & 54.25 
        & 28.53 & 28.54 & 13.42 \\
        
        & LSIMVC 
        & 32.31 & 60.67 & 17.26  
        & 92.92 & 86.21 & 85.11  
        & 21.92 & 24.06 & 7.34  
        & $\underline{86.38}$ & $\underline{77.12}$ & $\underline{68.83}$  
        & 34.65 & 38.35 & 19.56 \\ 
        
        & PIMVC  
        & $\underline{61.60}$ & $\underline{79.47}$ & $\underline{45.39}$  
        & $\underline{93.67}$ & $\underline{87.51}$ & $\underline{86.48}$
        & 19.35 & 25.03 & 6.64  
        & 79.14 & 69.88 & 60.68  
        & 34.82 & 36.72 & 18.66 \\ 
        
        & Ours  
        & $\bm{70.46}$ & $\bm{82.43}$ & $\bm{56.29}$  
        & $\bm{94.87}$ & $\bm{89.13}$ & $\bm{89.08}$  
        & $\bm{25.62}$ & $\bm{28.86}$ & $\underline{12.16}$  
        & $\bm{91.14}$ & $\bm{82.70}$ & $\bm{80.61}$  
        & $\bm{40.92}$ & $\bm{42.28}$ & $\bm{24.64}$ \\

        \hline
        \end{tabular}
\caption{The clustering results of nine methods on five complete datasets and incomplete datasets with the missing rate $\eta=0.5$, the best and second-best results are marked in bold and underlined, respectively.}
\label{table:evalution}
\end{table*}

\subsection{Compared Methods}
To verify the effectiveness and superiority of our approach, eight competing incomplete multi-view clustering methods are adopted to compare with our approach. These methods are described  as follows:
\begin{itemize}
\setlength{\itemsep}{0pt}
    \item \textbf{BSV}~\cite{zhao2016incomplete}:    
    It uses the average feature value to fill the missing data position and applies the K-means clustering algorithm to each view independently, selecting the best one as the final result.
    \item  \textbf{Concat}~\cite{zhao2016incomplete}: 
    It takes the same method of imputing missing values as BSV method, concatenates all of the views and then uses the K-means clustering method to obtain the result.
    \item  \textbf{DAIMC}~\cite{hu2019doubly}: 
    It employs weighted non-negative matrix factorization with a missing indicator matrix to derive a consensus representation.
    \item  \textbf{UEAF}~\cite{wen2019unified}:
    It integrates the missing view recovery and consensus representation learning with embedding alignment into a unified framework.
    \item  \textbf{ANIMC}~\cite{fang2021animc}:
    It automatically assigns appropriate weights to each view and uses double soft regularized regression to align the samples.
    \item  \textbf{DCP}~\cite{lin2022dual}:
    It maximizes the mutual information to learn consensus representation and minimizes the conditional entropy to recover missing data.
    \item  \textbf{LSMVC}~\cite{liu2022localized}:
    It learns a sparse consensus representation by optimizing a multi-view matrix factorization model that is sparsely regularized.
    \item  \textbf{PIMVC}~\cite{deng2023projective}:
    It learns a common subspace representation and uses graph constraints to extract structural information.
\end{itemize}
For all the compared approaches, we adopt the hyper-parameter setting in their original papers. To avoid randomness, we run all the methods five times and the mean values are reported. 

\subsection{Experimental Results and Analysis}
Table~\ref{table:evalution} reports the average value of the clustering results with each method running five times. From Table~\ref{table:evalution}, we can observe that in most cases, PIMVC performs second best but worse than ours, and our method almost performs better than all the other compared methods on all the five datasets with all the three metrics in both complete and incomplete scenarios. 

\begin{figure*}[!ht]
    \centering
    \includegraphics[width=\linewidth]{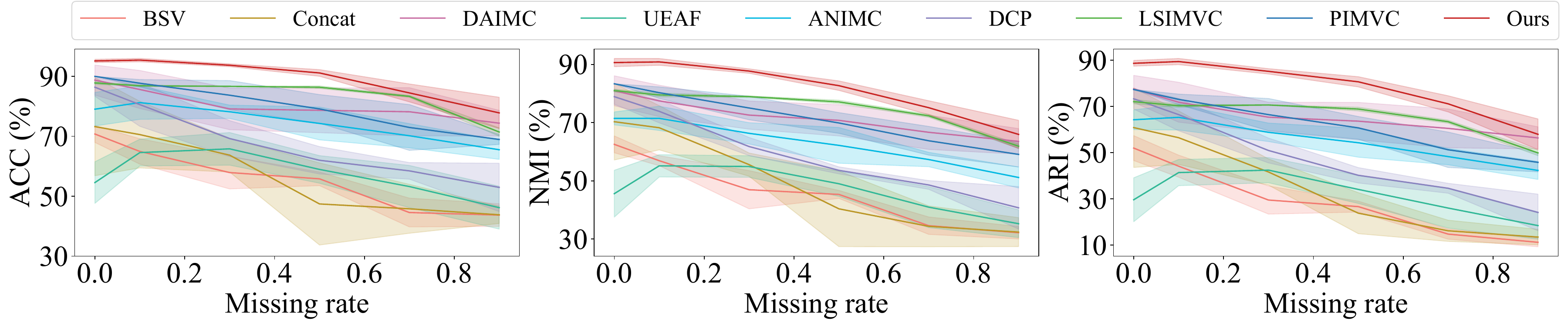}
    \caption{Comparison of the clustering performance on MSRC-V1 as missing rate increases, with padding representing the standard deviation.}
    \label{fig:error-band}
\end{figure*}

In order to validate the effectiveness of our model, we perform all the methods with the missing rate from 0.1 to 0.9 in intervals of 0.2.
Figure~\ref{fig:error-band} shows the performance of each model on MSRC-v1 dataset at different missing rates.
It can be seen that our model performs the best among all the methods on MSRC-v1 dataset with different missing rates, and our method still performs well even with high missing rates.

\begin{table}[!ht]
\begin{center}
\small
\begin{tabular}{c|ccc|ccc}
\hline
Missing Rate & $\mathcal{L}_{rec}$ &  $\mathcal{L}_{sc}$& $\mathcal{L}_{ccl}$ &  ACC & NMI & ARI \\ 
\hline
\multirow{4}{*}{\makecell{0.5}} 
&\ding{51}&\ding{51}&\ding{51}&  \textbf{91.14} & \textbf{82.70} & \textbf{80.61} \\
&\ding{51}&\ding{55}&\ding{51}& 87.81 & 79.48 & 75.64 \\
&\ding{55}&\ding{55}&\ding{51}& 86.00 & 78.34 & 74.51 \\
&\ding{51}&\ding{51}&\ding{55}& 67.62 & 59.23 & 41.39 \\
\hline
\end{tabular}
\end{center}
\caption{Ablation study on MSRC-V1 dataset with missing rate of 0.5.}
\label{table:ablation}
\end{table}

\subsection{Ablation Study}
To illustrate the effectiveness of each component of our model, we remove the view-specific representation learning module, global graph propagation module, and contrastive clustering module, respectively. Results of these cases on MSRC-v1 dataset with the missing rate of 0.5 are shown in Table~\ref{table:ablation}. When the global graph propagation module is removed, we only use view-specific GCN to impute representation. It should also be noted that when the view-specific representation learning module is removed, we cannot obtain the consensus representation, thus the global graph propagation module is removed together. We can see that the performance of the model would drop after deleting each module, which indicates the efficacy of each component. Especially the contrastive clustering module plays the most important role since the performance drops the most without it. The best performance is achieved when the three modules are considered simultaneously.

\begin{figure}[!ht]		
    \centering
    \subfloat[ACC]{
        \includegraphics[width=0.495\linewidth]{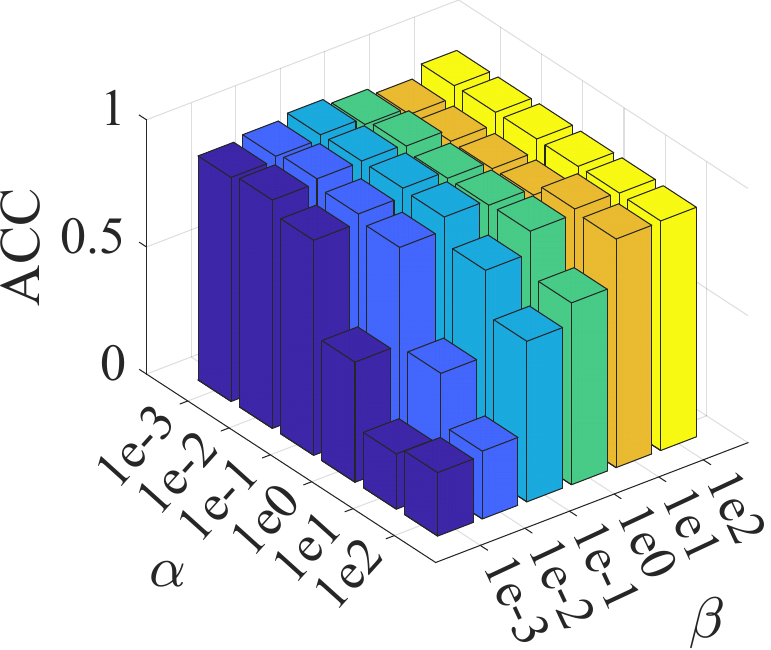}
    }
    \subfloat[NMI]{
        \includegraphics[width=0.495\linewidth]{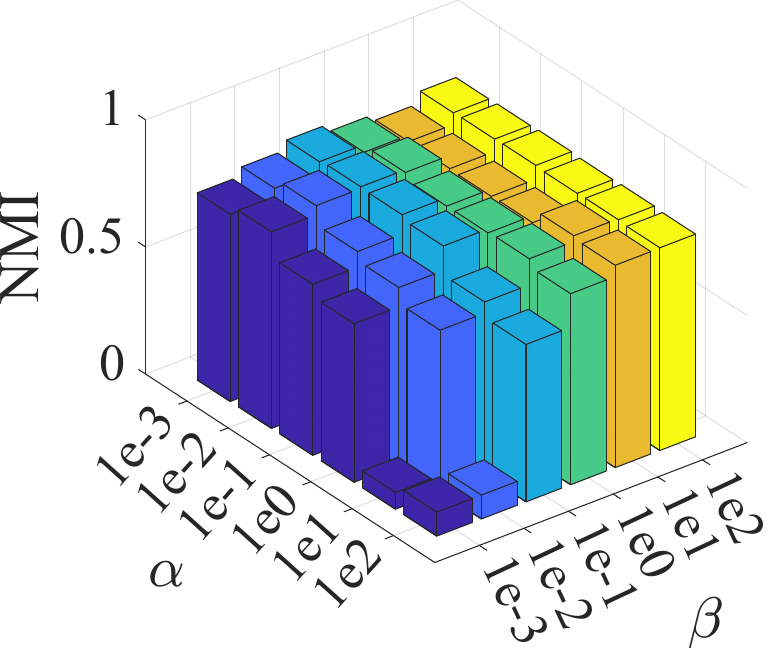}
    }
    \caption{The parameter sensitivity analysis of $\alpha$ and $\beta$ on MSRC-ataset with missing rate of 0.5.}
    \label{fig:hp-acc}	
\end{figure}

\begin{figure}[!ht]
    \centering
    \includegraphics[width=0.95\linewidth]{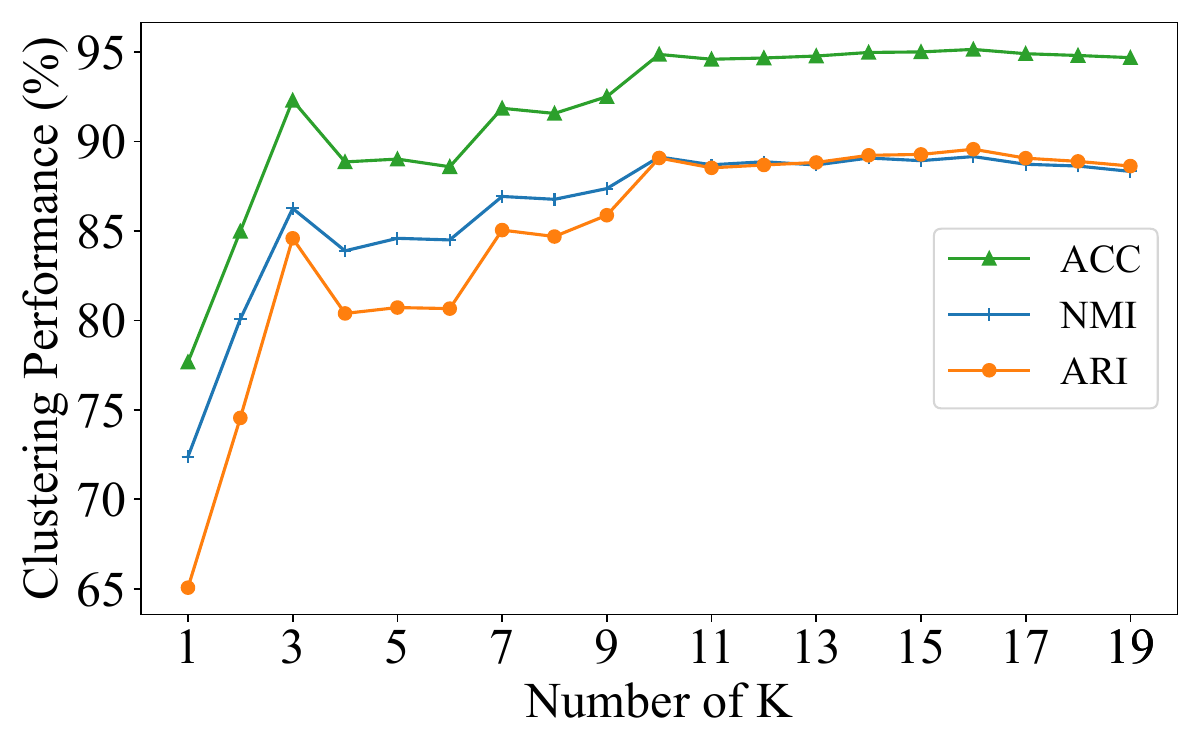}
    \caption{The influence of parameter $K$ on HandWritten dataset with missing rate of 0.5.}
    \label{fig:k-anl}
\end{figure}

\subsection{Parameter Sensitivity Analysis}
In this subsection, we analyze the parameter sensitivity of our model to parameters $\alpha$ and $\beta$. The two parameters vary from \{1e-3, 1e-2, 1e-1, 1e0, 1e1, 1e2\}. Figure~\ref{fig:hp-acc} illustrates the experiment results on HandWritten dataset. We can observe that our approach performs well except when $\alpha$ is much larger than $\beta$. These experimental results show that our method is insensitive to changes in parameters and can achieve great performance in a suitable parameter range.

We also conducted experiments on dataset HandWritten with the missing rate of 0.5 regarding parameter $K$ in the range [1, 20], as shown in Figure~\ref{fig:k-anl}. We can observe that small $K$ leads to poor performance as too few neighbors cannot reflect the structure well, while larger $K$ improves performance by capturing the correct graph structure.

\begin{figure}[!ht]		
    \centering
    \subfloat[missing rate of 0.1]{
        \includegraphics[width=0.49\linewidth]{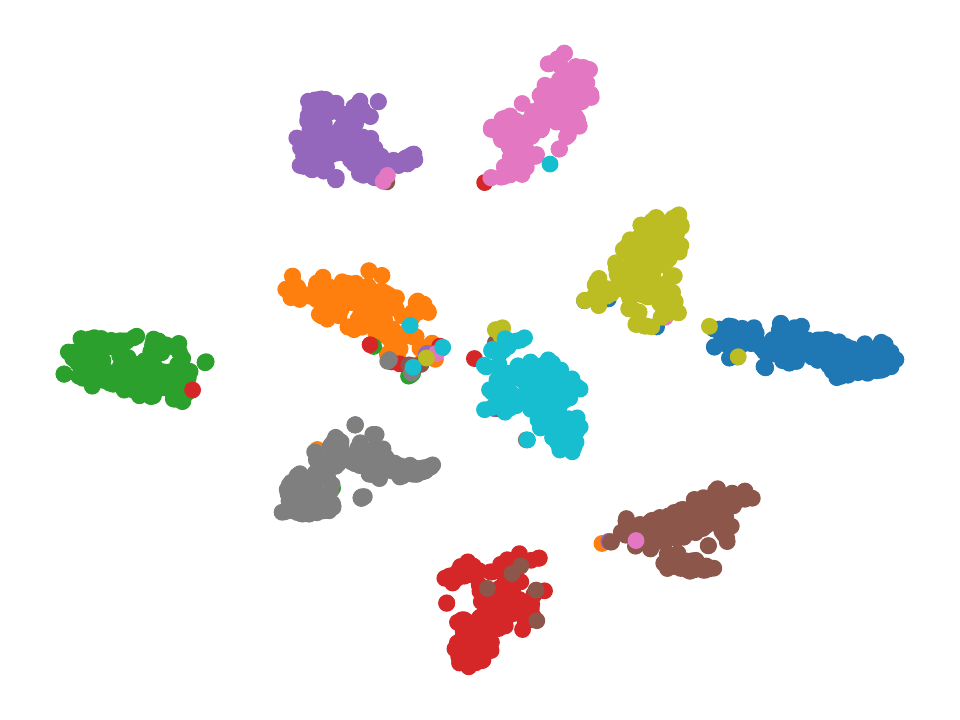}
    }
    \subfloat[missing rate of 0.3]{
        \includegraphics[width=0.49\linewidth]{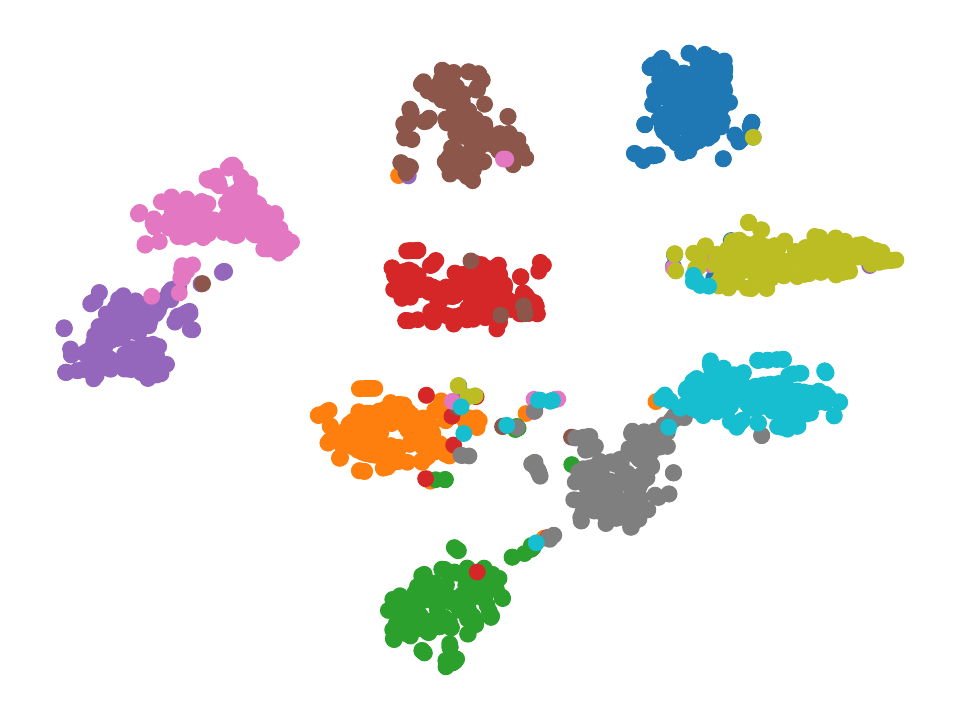}
    }  
    \quad
    \subfloat[missing rate of 0.5]{
        \includegraphics[width=0.49\linewidth]{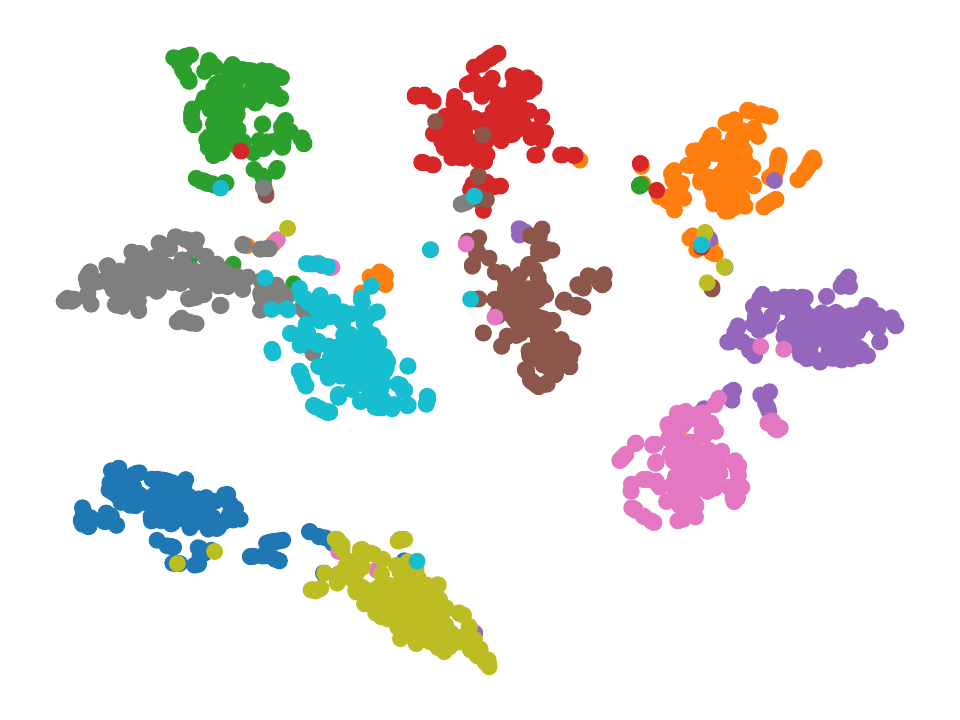}
    }  
    \subfloat[missing rate of 0.7]{
        \includegraphics[width=0.49\linewidth]{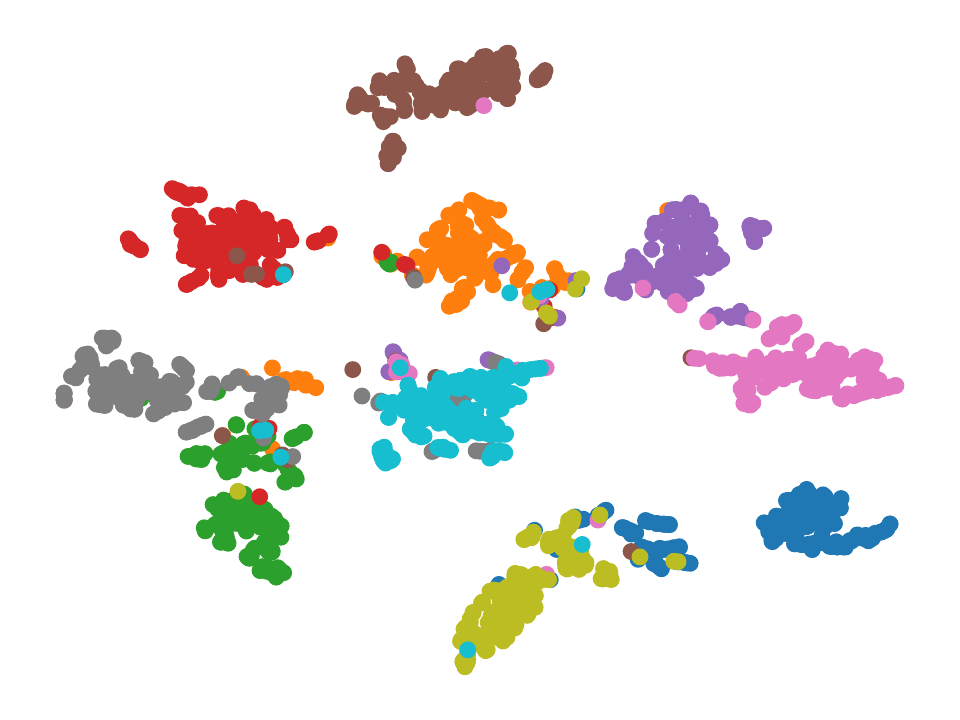}
    }
    \caption{The visualization results on HandWritten dataset with different missing rates.}
    \label{fig:vis-res}	
\end{figure}

\begin{figure}[!ht]
    \centering
    \includegraphics[width=0.99\linewidth]{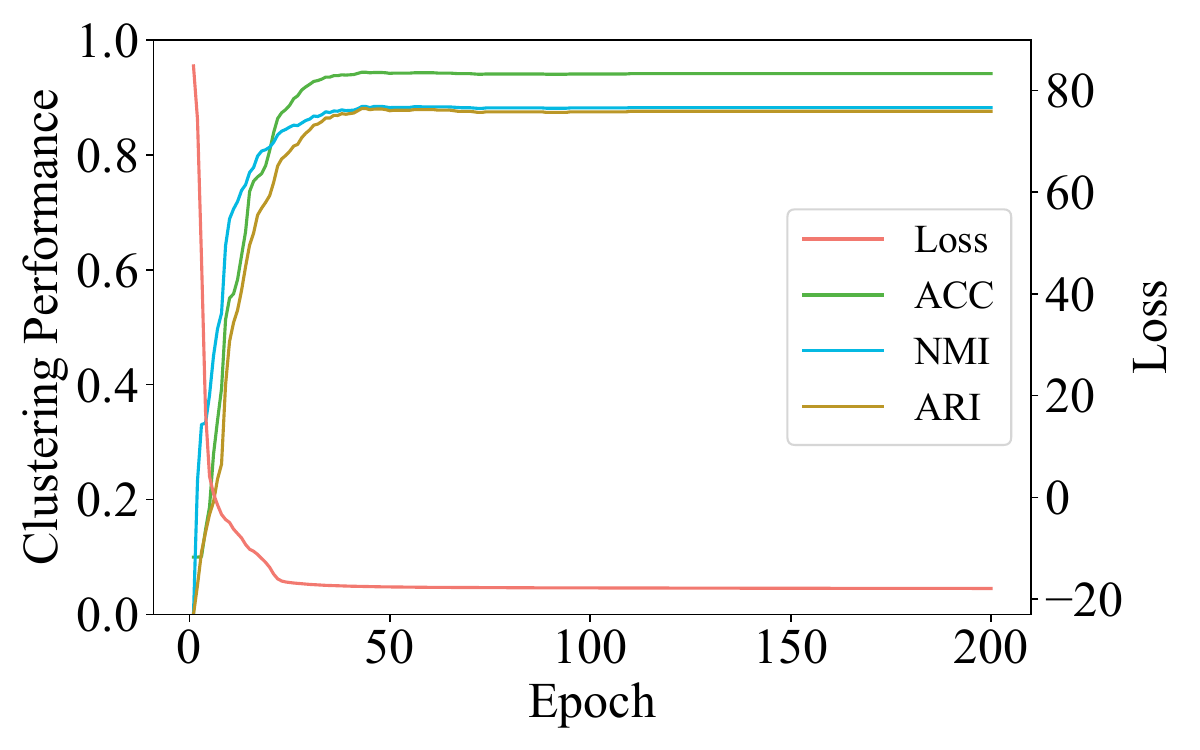}
    \caption{Illustration of the convergence analysis on HandWritten dataset with missing rate of 0.5.}
    \label{fig:loss}
\end{figure}

\subsection{Visualization Analysis}
We use t-SNE to visualize the clustering results on HandWritten dataset with different missing rates, as shown in Figure~\ref{fig:vis-res}. It can be observed that our model is able to separate the embeddings of the different categories effectively. Moreover, we noticed that with the increase of the missing rate, although the clustering results are not as well as when the missing rate is low, it still has a clear clustering structure.

\subsection{Convergence Analysis}
The convergence of our model on HandWritten dataset with a missing rate of 0.5 is illustrated in Figure~\ref{fig:loss}. As shown in Figure~\ref{fig:loss}, we can clearly observe that ACC, NMI, and ARI increase steadily and the loss decreases rapidly in the first few epochs. Afterward, as the number of epochs increases, ACC, NMI, and ARI grow gradually and finally converge to be stable, while the loss gradually declines with fluctuations.

\section{Conclusion}
In this paper, we propose an end-to-end framework named GHICMC for incomplete multi-view clustering. We have designed a global graph propagation module with hierarchical information transfer to handle the missing values adaptively, and conduct the contrastive learning on pseudo clustering assignments to obtain distinguishable clustering results. Extensive experiments conducted on five commonly used datasets verified the superiority of GHICMC over the state-of-the-art methods.

\section{Acknowledgments}
This work is supported in part by the National Natural Science Foundation of China (No. 62276079), Young Teacher Development Fund of Harbin Institute of Technology IDGA10002071, Research and Innovation Foundation of Harbin Institute of Technology IDGAZMZ00210325 and the Special Funding Program of Shandong Taishan Scholars Project.

\bibliography{aaai25}

\end{document}